\documentclass[10pt,twocolumn,letterpaper]{article}

\usepackage{wacv}              

\usepackage{graphicx}
\usepackage{amsmath}
\usepackage{amssymb}
\usepackage{booktabs}
\usepackage{array}
\usepackage{multicol}
\usepackage{multirow}
\usepackage{comment}
\usepackage{dblfloatfix}
\usepackage[accsupp]{axessibility}
\usepackage{nicefrac}

\usepackage[pagebackref,breaklinks,colorlinks]{hyperref}

\def\method{TakuNet} 

\usepackage[capitalize]{cleveref}
\crefname{section}{Sec.}{Secs.}
\Crefname{section}{Section}{Sections}
\Crefname{table}{Table}{Tables}
\crefname{table}{Tab.}{Tabs.}

\def\wacvPaperID{7} 
\def\confName{WACV}
\def\confYear{2025}

\begin{document}

\title{\method: an Energy-Efficient CNN for Real-Time Inference \\ on Embedded UAV systems in Emergency Response Scenarios}

\author{Daniel Rossi, Guido Borghi, Roberto Vezzani\\
University of Modena and Reggio Emilia, Italy \\
{\tt\small \{name.surname\}@unimore.it}
}
\maketitle

\begin{abstract}
Designing efficient neural networks for embedded devices is a critical challenge, particularly in applications requiring real-time performance, such as aerial imaging with drones and UAVs for emergency responses. In this work, we introduce \method, a novel light-weight architecture which employs techniques such as depth-wise convolutions and an early downsampling stem to reduce computational complexity while maintaining high accuracy. It leverages dense connections for fast convergence during training and uses 16-bit floating-point precision for optimization on embedded hardware accelerators. Experimental evaluation on two public datasets shows that \method\ achieves near-state-of-the-art accuracy in classifying aerial images of emergency situations, despite its minimal parameter count. Real-world tests on embedded devices, namely Jetson Orin Nano and Raspberry Pi, confirm \method's efficiency, achieving more than 650 fps on the 15W Jetson board, making it suitable for real-time AI processing on resource-constrained platforms and advancing the applicability of drones in emergency scenarios. The code and implementation details are publicly released\footnote{\url{https://github.com/DanielRossi1/TakuNet}}.
\end{abstract}
\section{Introduction}
\label{sec:Introduction}

Images captured at high altitudes are very important data that, especially if processed by artificial intelligent systems, can provide crucial information in many areas, ranging from precision agriculture~\cite{daponte2019review}, environmental and urban monitoring~\cite{gallacher2016drone}, energy resource planning~\cite{day2017drones}, to search and rescue operations~\cite{dousai2021detection} and risk prevention~\cite{furutani2021drones}. 

Usually, high-altitude images are acquired by satellites, aircraft, or even drones and unmanned aerial vehicles (UAVs), where the latter has begun to gain increasing interest since they combine affordability and the acquisition of high-quality and stabilized images. In addition, in emergency response scenarios, drones and UAVs play a key and crucial role, such as entering remote or difficult-to-access areas, providing a comprehensive view of the hazard, and enabling safer conditions for the personnel engaged in rescue operations.

Unfortunately, the adoption of modern artificial intelligence systems in this kind of device is still critical and poses specific and significant challenges, mainly related to the limitation imposed by embedded hardware~\cite{mersheeva2015multi}. 
Specifically, the power consumption of drone and UAV hardware is constrained by the use of lightweight batteries, which limits the feasibility of employing high-frequency processors.
Therefore, in the context of embedded systems, an efficient design is a key feature, especially for deep neural networks~\cite{oh2017investigation}, such as CNNs and Transformers. 

In this context, a possible solution is to run neural network architectures in the cloud. Images or videos can be sent to a remote server that takes care of processing them. 
We observe that this approach is not always feasible due to, for instance, damages to telecommunication infrastructure or lack of signal in remote areas. 
Another solution is represented by the recent introduction on the market of new ARM-based silicon architectures that combine increasingly high efficiency with ever-growing computational power~\cite{ryzhyk2006arm}. 
Finally, a complementary approach is represented by the development of efficient neural network architectures, \ie networks that require a limited amount of computing capabilities and memory, possibly maintaining a good level of accuracy~\cite{moon2020compression} with respect to standard models.

Following the latter approach, in this paper, we introduce \method, an extremely lightweight architecture explicitly designed to limit the number of parameters and, in general, the computational load and then foster its use on embedded boards. 
Indeed, the number of parameters in a model is a crucial value to consider when designing the architecture: a deficiency of parameters leads to a network that is unable to properly model and learn information from the data; on the other hand, too many parameters means wasting resources. Consequently, it is important to find the right balance, especially if the target hardware is a battery-powered embedded system with limited resources. 

From a technical point of view, \method\ is mainly based on depth-wise convolutions~\cite{howard2017mobilenets}, and exploits an early downsampling stem to reduce computational complexity in the initial layers. \method\ consists of four different stages, at the end of each a downsampler block reduces the size of the feature maps while implementing a dense connection which allows for fast convergence of the model during training. \method\ is trained with a floating-point resolution of 16 bits, enabling lossless optimization on hardware accelerators for embedded systems. Through the utilization of NVIDIA TensorRT\footnote{\url{https://developer.nvidia.com/tensorrt}}, a substantial enhancement in performance is achieved.

The experimental evaluation of model accuracy is conducted on two public datasets, namely AIDER~\cite{kyrkou2019deep} and AIDERv2~\cite{shianios2023benchmark}, both containing aerial images depicting emergency situations, \eg floods, traffic accidents, and fires. The proposed method achieves an accuracy close to the state-of-the-art, despite a very limited number of parameters.
Finally, we test \method\ on embedded devices, \ie Jetson Orin Nano and Raspberry Pi, demonstrating its real-time performance and potential for deployment on drones in emergency response scenarios.

In summary, the contributions of this work are the following ones:
\begin{enumerate}
    
    \item We introduce \method, a new architecture specifically designed to limit the number of parameters and computational complexity. This promotes the adoption of our architecture on embedded boards, \eg drones and UAVs, commonly employed in emergency scenarios. 

    \item We assess the performance of our model using the F1-score metric on two public datasets, attaining high accuracy despite its constrained computational requirements.

    \item We evaluate our architecture on four real-world embedded platforms, \ie Jetson Orin Nano and three Raspberry Pi. \method\ achieves more than $650$ fps on the Jetson board when optimized with TensorRT.
    
\end{enumerate}

\section{Related Work}
\label{sec:Related Work}
\subsection{Performance of Deep Learning architectures}
Convolutional architectures for image processing are among the most studied in the field of Computer Vision. 

The first notable architecture was AlexNet \cite{NIPS2012_c399862d},  the first deep model trained on a GPU, which used dropout layers and ReLU activation functions and won the 2012 ILSVRC competition.
Unlike the previous method, VGG \cite{simonyan2014very} promoted the use of small kernels in convolutional layers and a uniform architecture that progressively increased the number of channels in the feature maps.
As the networks gradually became deeper, the problem of vanishing gradient began to be noticed, making weights update negligible. ResNet \cite{he2016deep} demonstrated the effectiveness of skip connections, in which the output of a layer was added to its input, resulting in extremely deep architectures. 
Differently, DenseNet \cite{huang2017densely} exploited dense connections between the input and output of its blocks, promoting feature reuse and improving model efficiency. 

Unlike previous approaches, the introduction of Vision Transformers (ViT) \cite{dosovitskiy2020image, vaswani2017attention} for image processing significantly increased both the number of parameters and the overall complexity of the models. ViTs \cite{dosovitskiy2020image} can model global dependencies and consequently extract more relevant features for vision tasks. Unfortunately, the attention mechanism scales its complexity quadratically, limiting performance and easily saturating the memory due to the huge matrices that need to be computed.

In summary, despite being able to model only local relationships between feature maps, CNNs are fundamentally designed to be efficient through the use of shared kernels. Recent developments in CNNs have made it possible to further reduce their complexity while preserving or enhancing their output quality. 

\subsection{Towards more efficient architectures}
One of the pioneering approaches in neural network efficiency is SqueezeNet \cite{iandola2016squeezenet}, which reduces the size of the model by more than 500 times compared to AlexNet.
It employs ``fire modules'', which integrate ``squeeze''
and ``expand'' convolution layers, lowering the number of parameters while preserving high performance. 

Subsequently, ShuffleNet \cite{zhang2018shufflenet} improves efficiency by introducing the ``shuffle'' operation, which rearranges feature maps to enhance communication between convolution groups. This overcomes the limitations of group-wise convolutions that restrict interaction along the channel axis.

One of the most efficient convolutional blocks was introduced in MobileNetV1 \cite{howard2017mobilenets}, namely depthwise separable convolution. 
They consist of a depthwise convolution applied independently to each input channel, followed by a $1\times1$ convolution that combines the resulting feature maps along the channels, leading to a reduction in parameters and computational costs compared to traditional 2D convolutions.
MobileNetV2 \cite{sandler2018mobilenetv2} enhances accuracy through the introduction of the ``inverted residual structure'', which first expands the channel size using pointwise convolutions before applying depthwise convolutions, in contrast to traditional architectures. Additionally, it incorporates shortcut connections between bottleneck blocks, akin to ResNet networks, facilitating direct information flow, improving gradient propagation, and aiding in network training.
Later, MobileNetV3 \cite{howard2019searching} optimizes its architecture using Neural Architecture Search (NAS) to enhance design efficiency. It simplifies the model by removing certain layers and introduces the HardSwish activation function. Additionally, MobileNetV3 incorporates ``squeeze-and-excitation'' \cite{hu2018squeeze} blocks, allowing the network to recalibrate channel activations, emphasizing informative features while suppressing less relevant ones, thereby improving the quality of learned representations.

EfficientNet \cite{tan2019efficientnet} introduces a novel approach to scaling neural networks through compound scaling, which uniformly adjusts the network's width, depth, and resolution to optimize performance. This method allows for efficient adaptation of larger models to smaller datasets and vice versa, ensuring high output quality while enhancing overall efficiency.

In addition to solutions based on efficient architectural choices, neural networks can also be optimized a posteriori. Pruning and quantization~\cite{Vorabbi202425} are two of the most well-known strategies used to optimize neural networks. Pruning works by removing unnecessary weights or connections from a model, which reduces its size and computational demands. This can lead to faster inference times and a smaller memory footprint, while often preserving accuracy if done correctly. However, pruning can be complex to implement and its improper execution may result in a loss of model performance.
Quantization, on the other hand, involves converting the model's weights and activations from floating-point precision to lower-bit representations, such as 8-bit integers. This approach improves computational efficiency and reduces energy consumption.
Despite these benefits, quantization, if not carefully managed, carries the risk of accuracy degradation. Besides, not all hardware supports low-precision computations, limiting its applicability.

\subsection{Challenges in aerial images recognition tasks}
In the context of image recognition, datasets containing frames captured at high altitudes pose different challenges than scenes captured on the ground. 
For instance, many aerial data collections~\cite{helber2019eurosat, WULDER2016271, toker2022dynamicearthnet} contain multispectral imagery, which requires appropriate extraction or fusion layers of the spectral bands of input features. The various spectral feature maps can be composed of images captured in the visible light band, infrared, ultraviolet, and so on. Sensors for multiband acquisition have exorbitant costs and are often mounted on satellites, making updating such datasets an expensive and burdensome operation. 

In addition, real-time monitoring through satellite imagery is challenging, as satellites take hours, if not days, to sample specific points on the planet multiple times. However, images acquired from drones or aircraft hardly integrate multispectral images. Nevertheless, these devices allow to perform real-time monitoring of specific and even remote areas, providing access to data at a lower latency. 

Regardless of the acquisition system, high-altitude imaging is characterized by the fact that objects are very small and details are quite difficult to capture. Aerial datasets such as \cite{kyrkou2019deep, shianios2023benchmark} provide a more extensive overview of the area being monitored and find easy use in the analysis and recognition of, for example, violent weather events, such as forest fires, floods, major vehicle accidents, or collapse of buildings.

\begin{figure*}[h!]
    \centering
    \includegraphics[width=1\textwidth]{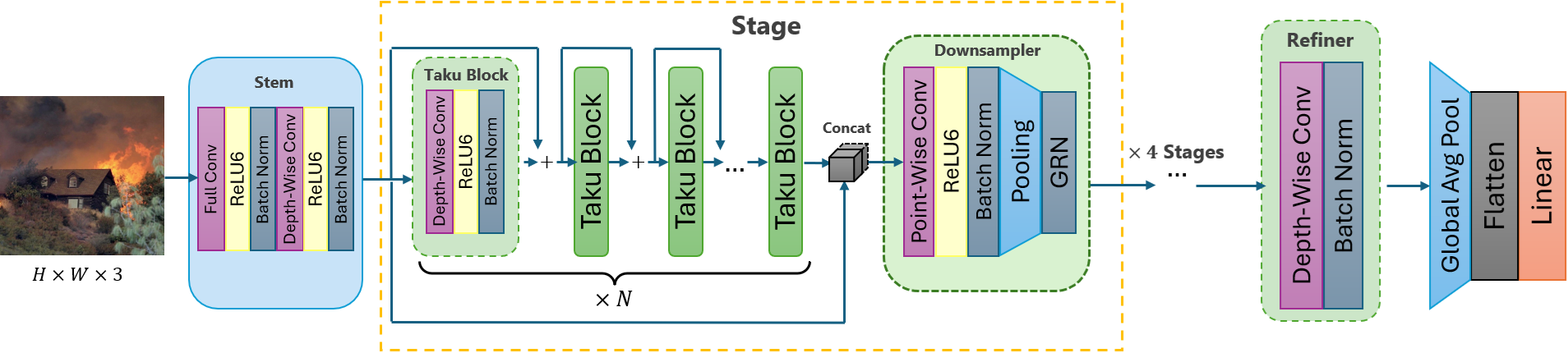}
    \caption{Overview of the \method\ architecture. The input image is first processed by a convolutional stem block. Four subsequent stages progressively extract spatial features, whose final output is later integrated along the channel axis with dense connection feature maps coming from the stage's input. At the end of each stage, the Downsampler block reduces the spatial size, while expanding the channel dimension. Finally, the Refiner block balances spatial features before the linear classification layer. }
    \label{fig:overview}
\end{figure*}

\subsection{Methods for aerial image recognition}
In recent years, aerial image classification has played a crucial role in various scenarios, including agriculture, where the ability to analyze and interpret visual data can significantly improve resource management. 

Yang \etal \cite{rs9060583} approach combines spatial and spectral information through a hybrid classification technique, achieving high accuracy in classifying damaged rice lodging areas using drone-acquired imagery. Mafania \etal \cite{MAFANYA20171} proposes to map the invasive plant Harrisia pomanensis using UAV imagery, comparing different image classifiers to assess their effectiveness in mapping this species. Among the later methods for aerial image classification, APDC-Net proposed by Bi \etal \cite{bi2019apdc} is distinguished by the use of attention-based pooling and a dense connection structure, which preserves multilevel features and improves local semantic representation. To improve object detection in UAV images, Hendria \etal \cite{HENDRIA2023258} proposed a combined approach that integrates transformer-based models \cite{liu2021swin} and CNN \cite{qiao2021detectors}, using ensemble and hybrid techniques. However, the shortcoming of these methods is that they are all designed to run offline and therefore do not allow for real-time processing of aerial images.
In this regard, two important contributions are introduced in \cite{kyrkou2020emergencynet} and \cite{tinyemergencynet}, who laid the groundwork for architectures capable of running on UAV(s). Kyrkou \etal introduced EmergencyNet \cite{kyrkou2020emergencynet}, a model based on atrous convolution, whose multi-resolution features are later merged. Mogaka \etal, derived TinyEmergencyNet \cite{tinyemergencynet} from the previous method by using a pruning strategy, removing up to $60\%$ of the least significant weights of the model. 

However, despite significant advancements in efficient neural architectures, many models remain too complex for widespread deployment on embedded devices. There is a clear gap in approaches specifically tailored for these systems, particularly regarding micro-components like activation functions, which can exhibit varying latency. Furthermore, existing hardware accelerators for neural networks are often overlooked in architecture design. TakuNet aims to address these challenges by consolidating and harmonizing existing techniques that enhance neural network efficiency, focusing on minimizing parameters and latency specifically for embedded devices. It is specifically designed to leverage available hardware accelerators, ensuring that the weights learned during training remain intact.

\section{\method\ model}
\label{sec:Method}
The architecture of the proposed \method\ is depicted in Figure~\ref{fig:overview}. 
Our choices are driven by the need to develop a model that combines strong performance in image classification with limited computational demands -- \textit{i.e.} a limited number of parameters and floating point operations -- enabling real-time execution on embedded hardware.

From a general point of view, we take advantage of several elements introduced in the recent literature. The first is the depth-wise convolutions and grouped point-wise convolutions, inspired by~\cite{howard2017mobilenets, schuler2022grouped} to reduce the number of convolutional layer parameters. 
We harness the Global Response Normalization layer~\cite{woo2023convnext} to promote feature diversity and increase contrast and selectivity along channels, enabling the model to identify and differentiate more effectively between relevant features. 
In addition, dense connections~\cite{huang2017densely} are used to encourage feature reuse and better gradient propagation.  
Finally, following ~\cite{kyrkou2020emergencynet}, we have included a layer of dilated convolution in the stem to capture more distant features without enlarging the number of parameters. 

\subsection{\method\ Architecture}
From an architectural point of view, \method\ consists of $6$ main macroblocks -- \ie a stem block, $4$ stages containing Taku Blocks and a Downsampler block, a Refiner block, and a final linear layer that acts as a classifier. 

The first block is the stem, consisting of two stacked convolutional layers according to the following formulation:

\begin{equation}
\begin{aligned}
    y &= \mathrm{ReLU6}(\mathrm{BN}(\mathrm{Conv}_{3 \times 3}(x))), \\
    z &= \mathrm{ReLU6}(\mathrm{BN}(\mathrm{DWConv}_{3 \times 3}(y))) + y,
\end{aligned}
\end{equation} \\
where in this formula, and in the next sections of the paper, ReLU6 and BN denote Rectified Linear Unit function \cite{howard2017mobilenets}  clamped between $[-\infty, \,6]$, and Batch Normalization \cite{ioffe2015batch}, while $\mathrm{DWConv}$ refers to depthwise convolution.

The initial $3 \times 3$ full-convolution has a stride of $2$, a padding of $2$, and a dilation of $2$. An additional $3 \times 3$ depth-wise convolution layer follows, with stride $2$ and padding $2$. In these early layers of the network, the size of the initial image is reduced from $H\times W\times 3$ to $\frac{H}{4} \times \frac{W}{4} \times 40$. This plays a crucial role in significantly reducing computational complexity, as each successive convolutional layer will need to be applied fewer times on the feature maps it receives as input.

\begin{figure*}[th!]
    \centering
    \includegraphics[width=0.95\textwidth]{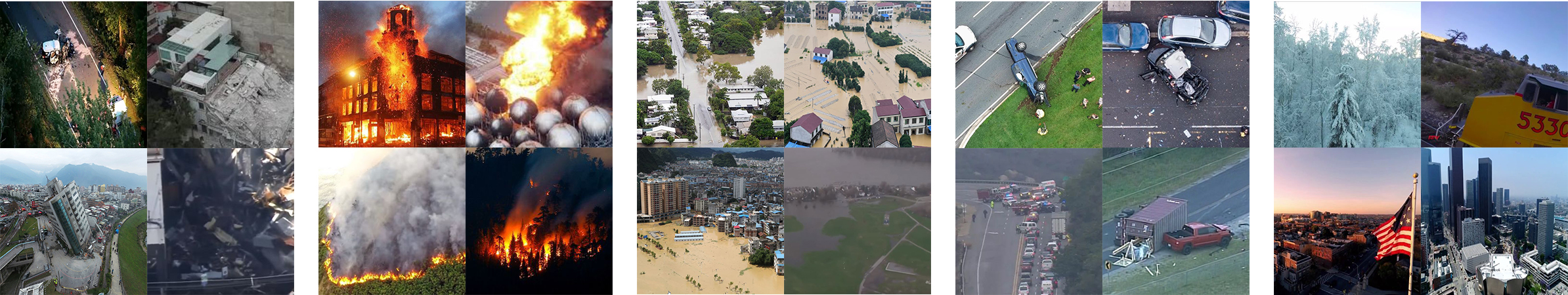}
    \caption{The variety of some of the images in the AIDER \cite{kyrkou2019deep} dataset grouped by each class. In detail, from the left, collapsed buildings, fire/smoke, floods, traffic incidents and normal classes are shown.}
    \label{fig:label}
\end{figure*}

The architecture subsequently consists of 4 stages. Each stage is composed of multiple Taku Blocks, which can be represented by the following formulation:
\begin{equation}
\begin{aligned}
    y &= \mathrm{ReLU6}(\mathrm{BN}(\mathrm{DWConv}_{3 \times 3}(x))) + x, \\
\end{aligned}
\end{equation}
in which a depth-wise convolution with stride $1$ is applied. The dilation and padding are set to $1$ for all Taku Block of each stage. The four stages have a depth of $(5,5,5,4)$, respectively. The number of stages and blocks are empirically set, aiming for the best tradeoff between accuracy on one side and model size and efficiency on the other side. At the end of each stage, a Downsampler block is responsible for reducing the size of the feature maps, while expanding the number of channels. Inspired by \cite{huang2017densely,zhang2018shufflenet}, the feature maps produced by the last Taku Block are concatenated and mixed with those coming from the dense connection. To further reduce the number of parameters, we apply a grouped point-wise convolutional layer \cite{schuler2022grouped} on the new concatenated feature maps, where the number of groups is given by $\left\lfloor\nicefrac{(\text{in\_channels} + \text{out\_channels})}{4} \right\rfloor$. Here $in\_channels$ refers to the number of channels of the feature maps input to the stage, while $out\_channels$ refers to the number of channels of the feature maps output to the last Taku Block. 

Next, a Batch Normalization and a ReLU6 activation function are applied. In the first 3 stages of the architecture, the output is then dimensionally scaled by a max-pooling layer, while in the last stage average pooling is applied. Each of these pooling layers is then followed by a Global Response Normalization (GRN) \cite{woo2023convnext} function.

Following the last stage, a Refiner block is applied to the feature maps according to the following formulation:
\begin{equation}
\begin{aligned}
    y &= \mathrm{AdaptiveAvgPool}(\mathrm{BN}(\mathrm{DWConv}_{3 \times 3}(x))), \\
\end{aligned}
\end{equation}
where the adaptive average pooling layer reduces the height and width dimensions of the feature maps to a unit value. The probability of each class is derived through a linear layer applied to the latent vector generated by the Refiner.

\section{Experiments}
\subsection{Aerial Image Disaster Event Recognition (AIDER) dataset}
The AIDER \cite{kyrkou2019deep} dataset is a collection of images acquired by aerial vehicles coming from various sources, such as the world-wide-web (YouTube, newspaper web pages, the image section of various search engines, etc.) and via real-world drone acquisitions. The dataset consists of a set of multi-resolution images of complex scenes composed of natural territories, urban environments, large road infrastructure, rural or coastal areas, and so on. It includes scenes acquired in different seasons and at distinct times of the day, trying to offer a truthful overview of the real world, despite its small size. The hand-collected images contain scenes of emergency situations such as floods, collapsed buildings, traffic accidents, and fires. 
The first version of the dataset features 320 images for fires/smoke, 370 images for floods, 320 images for collapsed buildings/rubble, 335 images for traffic accidents, and 1200 images for the normal class, \ie aerial images free of the various types of disasters. 
The dataset has a first limited version in size and was later expanded in \cite{kyrkou2020emergencynet}. 
Unfortunately, this latest version has been reduced due to copyright infringement and a rather high number of images have been removed. The image composition of the AIDER dataset is detailed in Table \ref{tab:AIDER_class_distribution}, which enumerates the total number of images available along with a breakdown of the distribution between individual classes. 
For a fair comparison, we split the dataset into training and testing sets, using a proportion of $70/30$ for all classes except Normal, split with a proportion of $65/35$, keeping the same test proportions as \cite{tinyemergencynet} and \cite{kyrkou2020emergencynet}. 

Due to the scarcity of data, we did not leave room for a standalone validation set, which we replaced with a k-fold cross-validation method performed on the training set, with $k=5$.

\begin{table}[h!]
\centering
\setlength{\tabcolsep}{3pt} 
\renewcommand{\arraystretch}{1.1} 
\begin{tabular}{l c c c}
\toprule
\textbf{Class} & \textbf{Train} & \textbf{Test} & \textbf{Total} \\ \midrule
Collapsed Building & 395 & 146 & 511 \\ 
Fire/Smoke & 403 & 148 & 521 \\ 
Flood & 406 & 150 & 526 \\ 
Traffic Accidents & 376 & 139 & 485 \\ 
Normal & 3,250 & 1,540 & 4,390 \\ \midrule
\textbf{Total Per Set} & \textbf{4,830} & \textbf{2,123} & \textbf{6,433} \\ \bottomrule
\end{tabular}
\caption{Class distribution across training and testing sets of AIDER dataset \cite{kyrkou2019deep} (latest available version).}
\label{tab:AIDER_class_distribution}
\end{table}

\subsection{Aerial Image Disaster Event Recognition Version 2 (AIDERv2) Dataset}
Similarly to AIDER, AIDERv2 \cite{shianios2023benchmark} is a collection of aerial images representing multiresolution emergency situations. The dataset collects urban, suburban, rural, coastal or lagoon scenes, and natural parks, taken at different times and seasons. 
Unlike the previous, the dataset is fully available and official training, validation, and test sets are provided, where the split ratio is set to $80/10/10$, respectively, as expressed in Table \ref{tab:AIDERV2_class_distribution}. 

\begin{table}[h!]
\centering
\setlength{\tabcolsep}{3pt} 
\renewcommand{\arraystretch}{1.1} 
\begin{tabular}{ l c c c c}
\toprule
\textbf{Class} & \textbf{Train} & \textbf{Validation} & \textbf{Test} & \textbf{Total} \\ \midrule
Earthquakes & 1,927 & 239 & 239 & 2,405 \\ 
Flood & 4,063 & 505 & 502 & 5,070 \\ 
Fire & 3,509 & 436 & 436 & 4,384 \\
Normal & 3,900 & 487 & 477 & 4,864 \\ \midrule 
\textbf{Total Per Set} & \textbf{13,399} & \textbf{1,670} & \textbf{1,654} & \textbf{16,723} \\ \bottomrule
\end{tabular}
\caption{AIDERv2 \cite{shianios2023benchmark}: distribution across training, validation, and testing sets.}
\label{tab:AIDERV2_class_distribution}
\end{table}

\begin{table*}[th!]
\centering
\begin{tabular}{l c c c c}
\toprule
\textbf{Model} & \textbf{Parameters} & \textbf{Model Size (MB)} & \textbf{F1-score} & \textbf{FLOPS} \\
\midrule
VGG16 \cite{simonyan2014very}& 14,840,133 & 59.36 & 0.601 & 17.62G \\
ResNet50 \cite{he2016deep}& 23,518,277 & 94.07 & 0.917 & 4.83G \\
SqueezeNet \cite{iandola2016squeezenet} & 737,989 & 2.95 & 0.890 & 845.38M \\
EfficientNet B0 \cite{tan2019efficientnet} & 4,013,953 & 16.06 & \textbf{0.950} & 479.68M \\
MobileNetV2 \cite{sandler2018mobilenetv2} & 2,230,277 & 8.92 & 0.930 & 371.82M \\
MobileNetV3 \cite{howard2019searching} & 4,208,437 & 16.83 & 0.915 & 265.02M \\
ShuffleNet \cite{zhang2018shufflenet} & 1,258,729 & 5.03 & 0.908 & 176.82M \\
EmergencyNet \cite{kyrkou2020emergencynet} & 90,963 & 0.360 & 0.936 & 77.34M \\
TinyEmergencyNet \cite{tinyemergencynet} & \underline{39,334} & \underline{0.160} & 0.895 & \underline{36.30M} \\ \midrule
\textbf{\method}$_{\text{FP}=16}$ & \textbf{37,685} & \textbf{0.15} & \underline{0.943} & \textbf{35.93M} \\ 
\textbf{\method}$_{\text{FP}=32}$ & \textbf{37,685} & \textbf{0.15} & 0.938 & \textbf{35.93M} \\
\bottomrule
\end{tabular}
\caption{Comparison of model parameters, memory footprint, F1-score, and FLOPs on the latest AIDER~\cite{kyrkou2019deep} version, where images are set to $240\times240$ pixels. Elements highlighted in \textbf{bold} refer to the best results, while those \underline{underlined} refer to the second-best results.}
\label{tab:AIDER_results}
\end{table*}

\subsection{Training}
In the early stage of the training pipeline, we apply image augmentation as a pre-processing step to enhance pattern generation and increase variance in the limited datasets. These augmentations are the same as those indicated in~\cite{kyrkou2020emergencynet} and include color shifting, blurring, geometric transformations (translations, rotations, mirroring), random image cropping, sharpening, shadowing, illumination variations, and zooming. Each transformation is applied with a randomly selected probability between 0.05 and 0.5, ensuring the model does not interpret random variations as intrinsic dataset properties.
Thereafter, all images are scaled to the range $[0, 1]$ and resized to a fixed dimension of $240\times240$ pixels for height and width.

Trainings are conducted on a system equipped with NVIDIA RTX 4070 Ti Super and Intel i7-12700F. \method\ is trained using the RMSProp \cite{tieleman2012rmsprop} optimizer with decay $0.9$, momentum $0.9$, and an initial learning rate of $1\times 10^{-3}$, along with a L$_2$\text{-}regularization term of $1 \times 10^{-5}$. Training is performed over 300 epochs with a batch size of 64, using a step learning rate scheduler to adjust the learning rate at each epoch, as defined by the following formula:
\begin{equation}
\begin{aligned}
    \eta_t=\eta_0 \times \gamma^{\left\lfloor \nicefrac{t}{\text{step\_size}} \right\rfloor}
 \\
\end{aligned}
\end{equation}
where $\eta_{t}$ represents the learning rate at epoch $t$, $\gamma$ is the multiplicative factor for the learning rate, and $step\_size$ is the period of learning rate decay. We use a $\gamma$ value of $0.975$ with a $step\_{size}$ of $2$. Finally, a cross-entropy loss function is used to measure the error in the probability estimates generated by the final linear layer.

\subsection{Evaluation and Performance Metrics}
Similarly to \cite{kyrkou2020emergencynet, tinyemergencynet}, to validate the accuracy of the method proposed here, we use the F1-score metric. This type of metric is very effective at measuring the quality of a model in the case where the dataset is unbalanced. 

As performance metrics, we use FLOP as the unit of measurement of computational complexity. FLOPs indicate the number of floating-point operations that a system can perform in one second. This measure has been criticized in literature \cite{tan2019mnasnet, wu2019fbnet} because it often does not reflect networks performance when executed on a real device. In addition, the efficiency of neural network components can vary significantly depending on the computational capabilities of the underlying hardware, particularly the CPU. To address this limitation, we provide a practical understanding of our model's performance, reporting the framerate obtained on real-world embedded boards. This assessment highlights the model's latency, offering a more explicit perspective on its suitability for deployment in real-world scenarios.

\begin{table*}[th!]
\centering
\small
\begin{tabular}{l c c c c}
\toprule
\textbf{Model} & \textbf{Parameters} & \textbf{Model Size (MB)} & \textbf{F1-score} & \textbf{FLOPS} \\
\midrule
ConvNext Tiny \cite{liu2022convnet} & 27,820,000 & 111.29 & 0.940 & 4.46G \\
GCVit XXtiny \cite{hatamizadeh2023global} & 11,480,000 & 45.93 & 0.932 & 1.94G\\
Vit Tiny \cite{dosovitskiy2020image}& 5,530,000 & 22.1 & 0.873 & 1.08G \\
Convit Tiny \cite{d_Ascoli_2022} & 5,520,000 & 22.07 & 0.871 & 1.08G \\
MobileViT s \cite{mehta2021mobilevit} & 4,940,000 & 19.76 & 0.855 & 1.42G \\
MobileViT V2 0100 \cite{mehta2022separable} & 4,390,000 & 17.56 & 0.875 & 1.41G \\
EfficientNet-B0 \cite{tan2019efficientnet} & 4,010,000 & 16.05 & 0.862 & 0.41G \\
MnasNet \cite{tan2019mnasnet} & 3,110,000 & 12.43 & 0.897 & 0.34G \\
MobileNetV2 \cite{sandler2018mobilenetv2} & 2,230,000 & 8.92 & 0.893 & 0.33G \\
MobileViT xs \cite{mehta2021mobilevit} & 1,930,000 & 7.74 & 0.837 & 0.71G \\
ShuffleNet V2 \cite{ma2018shufflenet} & 1,260,000 & 5.03 & 0.852 & 0.15G \\
MobileVit V2 050 \cite{mehta2022separable} & 1,110,000 & 4.46 & 0.834 & 0.36G \\
MobileViT xxs \cite{mehta2021mobilevit} & 950,000 & 3.81 & 0.859 & 0.26G \\
MobileNetV3 Small \cite{howard2019searching} & 930,000 & 3.72 & 0.868 & 0.06G \\
DiRecNetV2 \cite{Shianios_2024} & 799,380 & 3.20 & \textbf{0.964}& 1.09G \\
SqueezeNet \cite{iandola2016squeezenet} & 730,000 & 2.90 & 0.845 & 0.26G \\
EmergencyNet \cite{kyrkou2020emergencynet} & 90,704 & 0.360 & 0.952 & 61.96M \\
TinyEmergencyNet \cite{tinyemergencynet} & \underline{39,075} & \underline{0.160} & 0.928 & \underline{31.57M} \\ \midrule
\textbf{\method}$_{\text{FP}=16}$ & \textbf{37,444} & \textbf{0.15} & \underline{0.958} & \textbf{31.38M} \\
\textbf{\method}$_{\text{FP}=32}$ & \textbf{37,444} & \textbf{0.15} & 0.954 & \textbf{31.38M} \\
\bottomrule
\end{tabular}
\caption{Comparison of Model Parameters, Size, F1-score, and FLOPs on AIDERv2 \cite{shianios2023benchmark}, where image height and width are set to $224$. Elements highlighted in \textbf{bold} refer to the best results, while those \underline{underlined} refer to the second-best results.}
\label{tab:AIDERV2_results}
\end{table*}

\section{Results}
To comprehensively evaluate our model in terms of key metrics, such as the number of parameters, memory footprint, FLOPs, and F1-score, we conduct a comparative analysis against several well-known convolutional architectures, with particular focus on EmergencyNet~\cite{kyrkou2020emergencynet} and TinyEmergencyNet~\cite{tinyemergencynet}. These architectures are chosen because of their specific design for the aerial image recognition task.

Both EmergencyNet and TinyEmergencyNet are re-implemented using the PyTorch\footnote{\url{https://pytorch.org}} framework, adhering rigorously to the instructions detailed in the respective articles \cite{kyrkou2020emergencynet, tinyemergencynet}. During this process, we observe minor discrepancies in the publicly available implementation of EmergencyNet, while no official implementation of TinyEmergencyNet is found online. Other architectures included in the comparison are sourced directly from the TorchVision library\footnote{\url{https://pytorch.org/vision/stable}}, ensuring consistency and reproducibility in benchmarking. All methodologies in comparison are trained for 300 epochs with a batch size of 64 on AIDER (Table \ref{tab:AIDER_results}) and AIDERv2 (Table \ref{tab:AIDERV2_results}) datasets, employing the training settings specified in the respective publications.

\begin{table*}[ht!]
    \centering
    \small
    \begin{tabular}{l l c l c c}
        \toprule
        \textbf{Device} & \textbf{Accelerator} & \textbf{TDP}  & \textbf{Model} & \textbf{No. Images} & \textbf{FPS} \\ 
        \midrule
        \multirow{3}{*}{Raspberry Pi 3} & \multirow{3}{*}{CPU ARM Cortex-A53} & \multirow{3}{*}{~5W} & EmergencyNet     & 1000 & 6.7  \\ 
        & & &  \textbf{TinyEmergencyNet} & \textbf{1000} & \textbf{10.8} \\
        & & & \method          & 1000 & 7.0  \\ 
        \midrule
        \multirow{3}{*}{Raspberry Pi 4} & \multirow{3}{*}{CPU ARM Cortex-A72} & \multirow{3}{*}{~10W} & EmergencyNet     & 2500 & 14.7 \\ 
        & & & \textbf{TinyEmergencyNet} & \textbf{2500} & \textbf{22.0} \\  
        & & & \method          & 2500 & 16.7 \\
        \midrule
        \multirow{3}{*}{Raspberry Pi 5} & \multirow{3}{*}{CPU ARM Cortex-A76} & \multirow{3}{*}{~10W} & EmergencyNet     & 2500 & 43.3 \\ 
        & & & TinyEmergencyNet & 2500 & 59.8 \\ 
        & & & \textbf{\method} & \textbf{2500} & \textbf{62.1} \\ 
        \midrule
        \multirow{4}{*}{Jetson Orin Nano} & \multirow{4}{*}{GPU Tegra (Ampere)} & \multirow{4}{*}{~15W} & EmergencyNet     & 2500 & 140.0 \\ 
        & & & TinyEmergencyNet & 2500 & 160.0 \\ 
        & & & \textbf{\method}$_\text{TDP=15W}$          & \textbf{2500} & \textbf{657.0} \\ 
        & & & \method$_\text{TDP=7W}$          & 2500 & 564.0 \\ 
        \bottomrule
    \end{tabular}
\caption{Model performance evaluation on embedded platforms, with fps calculated as the mean latency over multiple runs with batch size equal to 1 and image size of $240$ pixels. TakuNet’s fps on Jetson Orin are reported after TensorRT optimization.}
\label{tab:fps_performance}
\end{table*}

\subsection{Results on AIDER}
We first perform a set of experiments on AIDER \cite{kyrkou2019deep} to validate the effectiveness of our method against some of the well-known CNNs, along with EmergencyNet and TinyEmergencyNet. Leveraging the stem to reduce the size of the input images by a factor of 4 has significantly enhanced the efficiency of our approach. Although our method employs a sparse integration of features along the channel axis, dense channel features integration enables \method\ to perform proficiently on AIDER while considerably reducing its parameter count. The results shown in Table \ref{tab:AIDER_results} highlight that \method\ is capable of surpassing EmergencyNet in F1-score while using the minimum number of parameters and FLOPs among the various competitors.

These results point out that our method is the best in terms of the trade-off between F1-score/FLOPs, making a significant contribution to high-efficiency aerial image classification methods. Furthermore, when \method\ is trained using floating point 32 precision, the F1-score obtained on the test set is lower to a minor extent. This result is important for two other reasons. First, the difference in F1-score between the use of 16-bit floating-point precision (FP=16) and 32-bit floating-point precision (FP=32) is moderated. However, the efficiency gap between the two is substantial, primarily due to hardware acceleration, enabling the optimization of half-precision computation. Secondly, not all processors support hardware acceleration for 16-bit floating-point precision. In such cases, either lossless 32-bit floating-point weight casting or 16-bit floating-point emulation can be employed.

\subsection{Results on AIDERv2}
To evaluate the robustness of our method, we also trained it on AIDERv2 \cite{shianios2023benchmark}, which consists of a larger number of images to be classified. 

As shown in Table \ref{tab:AIDERV2_results}, \method\ is able to provide the second-best F1-score with fewer parameters and FLOPs compared to competitors. Therefore, despite having a very small number of parameters, our method demonstrates its ability to compete with much larger architectures on more complex scenarios. 

Deeper experiments on AIDERV2 showcase the effectiveness of a few architectural design choices. In fact, by removing dense connections the model does not converge during training, while by omitting GRN the overall F1-score decreases to $0.950$. The channel shuffle and the refiner block also are key elements for the high-quality outputs of \method\, since their exclusion causes the accuracy to drop at $0.956$ and $0.955$, respectively.

\subsection{Performance on Embedded Platforms}
To evaluate the inference efficiency of \method, we performed multiple runs on various embedded platforms to measure latency and later compute related frames per second (fps). The results are summarized in Table \ref{tab:fps_performance}. Notably, \method\ exhibits limitations in performance on earlier embedded systems such as Raspberry Pi 3 and Pi 4. One potential bottleneck is the GRN layer, where the computation of norms may not be fully optimized for these specific CPUs. Indeed, as shown in Table \ref{tab:raspberrypi4_act_latency}, activation functions that require complex mathematical computations demonstrate increased latency on the Raspberry Pi 4.

\begin{table}[ht!]
\centering
\small
\begin{tabular}{lll}
\textbf{ReLU} \cite{nair2010rectified} & \textbf{ReLU6} \cite{howard2017mobilenets} & 
\textbf{LReLU} \cite {maas2013rectifier} \\ \midrule 
30.08s & 30.84s & 30.87s \\ \\ 
\textbf{ELU} \cite{clevert2015fast} & \textbf{CELU} \cite{barron2017continuously} & \textbf{GELU} \cite{hendrycks2016gaussian}\\ \midrule 
168.56s & 169.27s & 401.49s 
\end{tabular}
\caption{Latency of common activation functions on a Raspberry Pi 4, measured during execution on a (10000, 100) tensor.}
\label{tab:raspberrypi4_act_latency}
\end{table}

In contrast, on more advanced hardware, such as the Raspberry Pi 5, \method\ outperforms competing approaches, effectively demonstrating its ability to take advantage of the improved CPU power. The key benefit of our approach becomes evident with hardware acceleration, particularly when utilizing the GPU on Jetson devices. Since our method is trained using half-precision floating-point weights, it can be further optimized via TensorRT without any degradation in weight precision. This enables \method\ to surpass TinyEmergencyNet by a $4.1\times$ speed increase.

To further demonstrate the efficiency of our method, we restrict the Jetson Orin power consumption to a thermal design power (TDP) set to 7W, achieving a $3.5\times$ performance improvement compared to TinyEmergencyNet operating at 15W. This significant reduction in power consumption highlights the suitability of \method\ for resource-constrained environments.

\section{Conclusion}
This research highlights the importance of designing neural network models with a clear consideration of the target context, particularly the embedded devices intended for inference deployment. As demonstrated, combining existing state-of-the-art solutions in a systematic manner enables the creation of highly efficient architectures.

Our proposed \method\ achieves significant results, providing high accuracy compared to competing approaches while minimizing the parameter count, FLOPs, and memory usage. This makes it a versatile solution capable of addressing a wide range of requirements in embedded applications, particularly in emergency response scenarios. Additionally, we emphasize the significant impact of hardware-specific characteristics on model performance, such as the varying latencies of embedded devices when processing specific model components. By focusing on hardware-aware design, we demonstrate that impressive performance can be achieved through hardware acceleration, and that models trained with FP16 precision can be highly competitive. Such an approach is advisable for edge device models.

Future directions for this work include exploring pruning and quantization techniques to further reduce model complexity. These advancements could enable real-time execution \method\ on microcontrollers, facilitating the integration of emergency response systems into miniaturized and ultralow-power devices.

{\small
\bibliographystyle{ieee_fullname}
\bibliography{egbib}
}
\end{document}